\documentclass[11pt,a4paper,notitlepage]{article}

\usepackage{amsmath}
\usepackage{amsfonts}
\usepackage{amssymb}
\usepackage{amsthm}
\usepackage{dsfont}
\usepackage{centernot}

\usepackage{tikz}
\usetikzlibrary{arrows,fit,positioning}
\usepackage{graphicx}
\usepackage{hyperref}
\usepackage{subcaption}
\usepackage{natbib}
\usepackage[a4paper,margin=2cm]{geometry}

\pgfarrowsdeclarecombine{ring}{ring}{}{}{o}{o}

\tikzset{
    >=stealth',
    punkt/.style={
           circle,
           rounded corners,
           draw=black, thick,
           text width=1em,
           minimum height=1em,
           text centered},
    observed/.style={
           circle,
           rounded corners,
           draw=black, thick,
           minimum width=2.3em,
           minimum height=2.3em,
           font=\footnotesize,
           text centered,
           fill=black!10!white
           },
     latent/.style={
           circle,
           rounded corners,
           draw=black, thick, dashed,
           minimum width=2.2em,
           minimum height=2.2em,
           font=\footnotesize,
           text centered
           },
    target/.style={
           circle,
           rounded corners,
           draw=black, thick,
           minimum width=2.2em,
           minimum height=2.2em,
           font=\footnotesize,
           text centered,
           fill=black!20!white,
           },
    observedrect/.style={
           rectangle,
           rounded corners,
           draw=black, thick,
           minimum width=6em,
           minimum height=2em,
           font=\footnotesize,
           text centered,
           fill=black!10!white
           },
    latentrect/.style={
           rectangle,
           rounded corners,
           draw=black, thick, dashed,
           minimum width=2.2em,
           minimum height=2.2em,
           font=\footnotesize,
           text centered
           },
     targetrect/.style={
           rectangle,
           rounded corners,
           draw=black, thick,
           minimum width=6em,
           minimum height=2em,
           font=\footnotesize,
           text centered,
           fill=black!20!white,
           },
     empty/.style={
           circle,
           rounded corners,
           minimum width=.5em,
           minimum height=.5em,
           font=\footnotesize,
           text centered,
           },
    pil/.style={
           o->,
           thick,
           shorten <=2pt,
           shorten >=2pt,},
    sh/.style={ shade, shading=axis, left color=red, right color=green,
    shading angle=45 }
}

\theoremstyle{plain}
\newtheorem{theorem}{Theorem}

\theoremstyle{definition}
\newtheorem{definition}[theorem]{Definition}

\newtheorem{example}{Example}

\newcommand{\bigo}[1]{\mathcal{O}\left( #1 \right)}
\newcommand{\quotes}[1]{``#1''}
\newcommand{\set}[1]{\left\{#1\right\}}
\newcommand{\eqn}[1]{\begin{align}#1\end{align}}
\newcommand{\eq}[1]{\begin{align*}#1\end{align*}}
\newcommand{\cf}[2]{{#1}^{#2}}
\renewcommand{\P}[1]{\operatorname{P}\left(#1\right)}

\newcommand{\ci}{\mathrel{\perp\mspace{-10mu}\perp}}
\newcommand{\nci}{\centernot{\ci}}
\newcommand{\parents}[1]{\operatorname{\mathbf{Pa}}(#1)}
\newcommand{\ind}[1]{\mathds{1}\!\!\set{#1}}

\title{A Primer on Causal Analysis}

\author{Finnian Lattimore and Cheng Soon Ong}

\date{5 June 2018} 

\begin{document}

\maketitle

\begin{abstract}
\noindent
We provide a conceptual map to navigate causal analysis problems.
Focusing on the case of discrete random variables, we consider
the case of causal effect estimation from observational data.
The presented approaches apply also to continuous variables,
but the issue of estimation becomes more complex.
We then introduce the four schools of thought for causal analysis
\footnote{Parts of this document are copied verbatim from \citet{Lattimore2018}.}.
\end{abstract}

\newpage

\section{Conceptual Map}

Causal inference is an intuitively seductive phrase, and its use is often clouded in
mystery.
This document provides a brief primer about the different kinds of problems that can be
considered under the umbrella of causal inference or causal analysis.
Causal inference is often contrasted with statistical or probabilistic inference,
as captured by the phrase ``correlation does not imply causation''.
Vaguely speaking causal inference is the study of adding an extra requirement
to the definition of conditional probability models, such that we can mathematically
express the idea of a cause.
For standard statistical estimation (without an additional causal assumption),
there are already challenges in estimating conditional independence from
data~\citep{agresti02catda,zhang18larskm}.
These challenges do not disappear when considering causal models.
It is worthwhile to also consider how we are planning to use the results of causal
analysis. According to \citet{pearl18why} there are
three levels on the ladder of causation.
\begin{description}
  \item[association] Seeing and observing the environment.\\
    Is the incidence of lung cancer higher among smokers?
  \item[intervention] Doing and intervening in the environment.\\
    How do we reduce lung cancer? What is the effect if we ban cigarettes?
  \item[counterfactuals] Imagining, restrospection, understanding the environment.\\
    What if I had not smoked for the last two years?
\end{description}

Before we introduce the four main mathematical models of causal analysis in
Section~\ref{sec:four-schools-causality}, we briefly discuss two other major
dimensions for categorising causal inference. The first dimension
(Section~\ref{sec:structure-vs-effect}) distinguishes
between inferring of the graph of causes and using the graph of causes to
infer the effect of an action. An analogous problem exists also for non-causal
statistical analysis, which distinguishes between the choice of probabilistic
model class and the parameters of the model.
The second dimension, which is unique to causal analysis, is the kind of data
that we have available (Section~\ref{sec:type-of-data}). We may wish to perform
causal inference based on observational data or data we obtain after performing
interventions.

Given the challenges of performing causal inference, it is worth pointing out
that in some sense causality is ``inevitable'' if we find two dependent random variables.
\paragraph{Reichenbach’s common cause principle}
If two random variables $X$ and $Y$ are statistically dependent ($X \nci Y$),
then there exists a third variable $Z$ that causally influences both~\citep{peters17elecif}.
As a special case, $Z$ may coincide with either $X$ or $Y$.
Furthermore, this variable $Z$ screens $X$ and $Y$ from each other in the sense
that given $Z$, they become independent, $X \ci Y | Z$.
In practice, several other reasons can give rise to dependence:
There may be unobserved causes, also known as selection bias; the discovery of dependence
from data might be false due to performing multiple hypothesis testing; there may
be a common time dependence, for example both variables are growing exponentially.

\subsection{Structure discovery vs effect estimation}
\label{sec:structure-vs-effect}

Analogous to the distinction between the design choice of a class of probabilistic models
and the computation of their maximum likelihood parameters, there are two types of
causal analysis problems. First we may wish to discover the graph of causes from
a set of data. Second given the graph, we may wish to the effect of intervening on a
node in the graph.
Another way of stating the distinction is to consider whether we are
thinking about forward or reverse causal effects~\citep{gelman13whywfc}.
The forward direction considers the problem of estimating the effect of a given
intervention, and is akin to estimating the causal strength of an edge in the graph.
The reverse direction considers the problem of estimating
the cause of a particular observation. The reverse direction causal analysis
is similar to causal graph discovery as one would need to estimate which other
random variable (from the set of all random variables) caused the effect.

\subsubsection{Discovering causal graphs}

Causal discovery is the attempt to learn the structure of the causal relationships
between variables, on which there is a rich literature,
see \citet{Spirtes2016} for a recent review.
We may be interested in discovering a causal graph from a set of known models.
This arises naturally if we assume free access to a large observational data set,
from which the Markov equivalence class can be found via causal discovery techniques.
Work on the problem of selecting experiments to discover the correct causal graph from
within a Markov equivalence class~\citep{eberhardt2005,eberhardt2010causal,hauser2014two,Hu2014}
could potentially be incorporated into a causal bandit algorithm.
In particular, \citet{Hu2014} show that only $\bigo{\log \log n}$ multi-variable interventions are required on average to recover a causal graph over $n$ variables once purely observational data is used to recover the ``essential graph''.

\subsubsection{Effect estimation}

Given a causal graph, estimating the strength of causality is similar to estimating
the effect size in standard regression analysis~\citep{gelman06datarm}.
As we will see in Section~\ref{sec:counterfactuals}, we can consider the problem
of counterfactual reasoning in terms of two distinct conditional probability estimation
tasks. It turns out that these tasks are related (one factual and the second counterfactual)
and tools from domain adaptation can be used for estimation~\citep{johansson16learci}.

\subsection{Observation vs Intervention}
\label{sec:type-of-data}

In contrast to standard statistical analysis based on observational data,
causal analysis benefits from data that
is collected after performing an intervention on the system of interest. One experimental
design that allows us to study causal effects is the randomised control trial.

\subsubsection{Learning from observational data}

The goal of causal inference is to learn the effect of taking an action. We can do this directly via experimental approaches, however any given agent only has a limited capacity to manipulate the world. We are generating and storing data on almost every aspect of our lives at an unprecedented rate. As we incorporate sensors and robotics into our cities, homes, cars, everyday products and even our bodies, the breadth and scale of this data will only increase. However, only a tiny fraction of this data will be generated in a controlled way with the specific goal of answering a single question. An agent that can only learn from data when it had explicit control (or perfect knowledge of) the process by which that data was generated will be severely limited. This makes it critical that we develop effective methods that enable us to predict the outcome of an intervention in some system by observing, rather than acting on it. This is the problem of observational causal inference. The key feature that distinguishes observational from interventional data is that the learning agent does not control the action about which they are trying to learn.

\subsubsection{Learning from interventions}

The previous section focused on aspects of the problem of estimating the likely effect of an intervention from data gathered prior to making the intervention. There is an obvious alternative. Instead of trying to infer the outcome of an intervention from passive observations, one can intervene and see what happens. There are three key differences between observing a system and explicitly intervening in it. First, we determine the nature of the intervention and thereby control the data points used to estimate causal effects. Selecting data points optimally for learning is the focus of the optimal experimental design literature within statistics \citep{pukelsheim2006optimal} and the active learning literature in machine learning \citep{settles2010active}. Secondly, explicitly choosing interventions yields a perfect model of the probability with which each action is selected, given any context, allowing control over confounding bias. Finally, when we are intervening in a system we typically care about the impact of our actions on the system in addition to optimising learning. For example, in a drug trial, assigning people a sub-optimal treatment has real world costs. This leads to a trade-off between exploiting the best known action so far and exploring alternative actions about which we are less certain. This exploration-exploitation trade-off lies at the heart of the field of reinforcement learning \citep{sutton1998reinforcement}.

\subsection{Randomised experiments}
\label{sec:randomized_experiment}
\begin{figure}
\centering
\begin{tikzpicture}[->,shorten >=0pt,shorten <=0pt,node distance=3em,thick, node/.style={observedrect}, lt/.style={latent}]
\node[node](1){Treatment};
\node[node, above left=of 1](2){Randomiser};
\node[node, right=of 1](3){Outcome};
\node[lt, above left=of 3](4){U};
\path[]
	(2) edge (1)
	(1) edge (3)
	(4) edge (3);
\end{tikzpicture}
\caption{causal network for a randomised experiment}
\label{fig:random_experiment_network}
\end{figure}
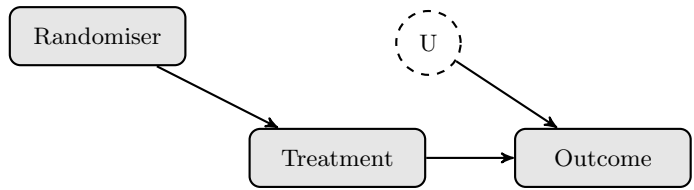

Randomised controlled trials are often presented as the gold standard for determining causal effects. What is it about randomisation that makes it so important when it comes to causality? The graphical model for a randomised controlled experiment is shown in figure \ref{fig:random_experiment_network}. If we assume perfect compliance (everyone takes the treatment that we select for them) then we have a perfect model for the treatment assignment process. Since treatment is assigned randomly, there can be no other variables that influence it and thus no confounding variables that affect both treatment and outcome.

Randomisation does not ensure target and control group are exactly alike. The more other features (observed or latent) influence the outcome, the more likely it is that there will be a significant difference in the joint distribution of these variables between the target and control groups in a finite data sample. However, the variance in the outcome, within both the target and control groups, also increases. The net result is increased variance (but not bias) in  the estimate of causal effects.

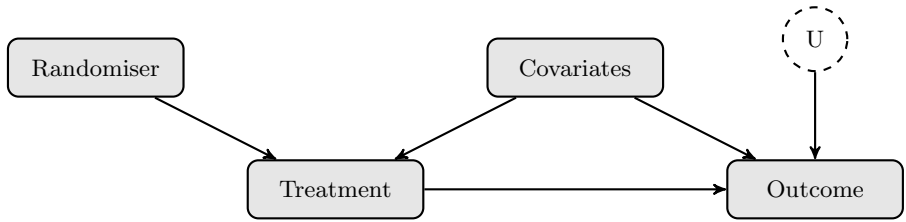
\begin{figure}
\centering
\begin{tikzpicture}[->,shorten >=0pt,shorten <=0pt,node distance=3em,thick, node/.style={observedrect}, lt/.style={latent}]
\node[node](1){Treatment};
\node[node, above left=of 1](2){Randomiser};
\node[node, above right=of 1](5){Covariates};
\node[node, below right=of 5](3){Outcome};
\node[lt, above=of 3](4){U};

\path[]
	(2) edge (1)
	(1) edge (3)
	(4) edge (3)
	(5) edge (3) edge (1);
\end{tikzpicture}
\caption{causal network for a stratified randomised experiment if the probability an individual is assigned a given treatment depends on some covariates.}
\label{fig:random_experiment_network_stratified}
\end{figure}

Stratified randomised experiments address the issue of variance due to covariate imbalance by randomly allocating treatment conditional on covariates believed to influence the outcome of interest. If we stratify in such a way that the probability an instance receives a given treatment is independent of its covariates, for example, by grouping instances by each assignment to the covariates and then assigning treatment randomly with fixed probabilities, the causal graphical model in figure \ref{fig:random_experiment_network} still holds. We can then estimate the average causal effects directly from the differences in outcome across treatments. More complex stratification strategies can introduce a backdoor path from treatment to outcome via the covariates on which treatment is stratified, (figure \ref{fig:random_experiment_network_stratified}). This necessitates that one condition on these covariates in computing the average causal effect in the same way as for estimating causal effects under ignorability. The key difference is that the propensity score is known, as it is designed by the experimenter, and there are guaranteed (rather than assumed) to be no latent confounding variables (that influence both treatment and outcome). See \citet{imbens2015causal} for a discussion of the trade-offs between stratified versus completely random experiments.

The benefit provided by randomisation in breaking the link between the treatment variable and any latent confounders should not be understated. The possibility of unobserved confounders cannot be empirically ruled out from observational data \citep{Pearl2000} (there is no test for confounding). This means causal estimates from non-experimental data are always subject to the criticism that an important confounder may have been overlooked or not properly adjusted for. However, randomised experiments do have some limitations.

\subsection{Limitations of randomised experiments}
\label{subsec:limitations_of_experiment}

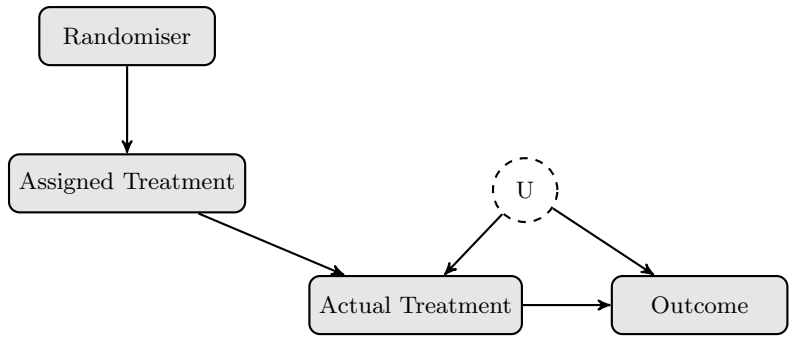
\begin{figure}
\centering
\begin{tikzpicture}[->,shorten >=0pt,shorten <=0pt,node distance=3em,thick, node/.style={observedrect}, lt/.style={latent}]
\node[node](1){Actual Treatment};
\node[node,above left=of 1](5){Assigned Treatment};
\node[node, above =of 5](2){Randomiser};
\node[node, right=of 1](3){Outcome};
\node[lt, above left=of 3](4){U};
\path[]
	(2) edge (5)
	(5) edge (1)
	(1) edge (3)
	(4) edge (3) edge (1);
\end{tikzpicture}
\caption{causal network for a randomised experiment with imperfect compliance}
\label{fig:random_experiment_network_imperfect_compliance}
\end{figure}

The idealised notion of an experiment represented by figure \ref{fig:random_experiment_network} does not capture the complexities of randomised experiments in practice. There may be imperfect compliance so that the treatment selected by the randomiser is not always followed, or output censoring in which the experimenter is not able to observe the outcome for all units (for example if people drop out). If compliance or attrition is not random, but associated with (potentially latent) variables that also affect the outcome, then the problem of confounding bias returns.\footnote{Non-compliance is a problem if the goal is to estimate the causal effect of the treatment on the outcome but not if the goal is to estimate the causal effect of prescribing the treatment. The latter makes sense in a context where the process by which people decide whether to take the treatment they have been prescribed is likely to be the same if the treatment were made available more generally beyond the experimental trial.} See figure \ref{fig:random_experiment_network_imperfect_compliance} for a graphical model of a randomised experiment with imperfect compliance.

It is not always possible or ethical to conduct a randomised controlled trial, as is beautifully demonstrated by the paper of \citet{smith2003parachute} on randomised cross-over trials of parachute use for the reduction of the mortality and morbidity associated with falls from large heights. When experimentation is possible, it is frequently difficult or expensive. This means experimental data sets are often much smaller than observational ones, limiting the complexity of models that can be explored. In addition, they are often conducted on a convenient, but unrepresentative, sample of the broader population of interest (for example first year university students). This can result in estimates with high \emph{internal validity} \citep{Campbell1963}, in that they should replicate well in a similar population, but low \emph{external validity} in that the results may not carry over to the general population of interest. The question of whether an experiment conducted on one population can be mapped to another is referred to as the transportability problem \citep{Bareinboim2013} and relies on very similar assumptions and arguments to causal inference and the do-calculus.

Finally, non-adaptive randomised experiments are not optimal from either an active
or reinforcement learning perspective. In fact it is becoming increasingly common
for adaptive clinical trials to be used as a study design~\citep{bothwell18adadct}.
As an experiment proceeds, information is obtained about the expectation and variance
of each intervention (or treatment). Fixed experimental designs cannot make use of this
information to select which intervention to try next.
This results in both poorer estimates for a fixed number of experimental samples
and more sub-optimal actions during the course of the experiment.

\section{Four schools of causality}
\label{sec:four-schools-causality}

In this section, we consider limit the discussion to the case where we already know
the causal graph and are only interested in estimating the causal effect from observational data.
To simplify the presentation we only consider discrete random variables with finite number
of states.

Observational causal inference aims to infer the outcome of an intervention in some system from data obtained by observing (but not intervening on) it. As previously mentioned, this is a form of transfer learning; we need to infer properties of the system post-intervention from observations of the system pre-intervention. Mapping properties from one system to another requires some assumptions about how these two systems are related, or in other words, a way of describing actions and how we anticipate a system will respond to them. Three key approaches have emerged: counterfactuals, structural equation models and causal Bayesian networks.

Counterfactuals \citep{Rubin1974} were developed from the starting point of generalising from randomised trials to less controlled settings. They describe causal effects in terms of differences between counterfactual variables, what would happen if we took one action versus what would happen if we took another. Counterfactual assertions can be expressed very naturally in human languages and are  prevalent in  everyday conversations; \quotes{if I had worked harder I would have got better grades} and \quotes{she would have been much sicker if she hadn't taken antibiotics}. Structural equation models have been developed and applied primarily within economics and related disciplines. They can be seen as an attempt to capture key aspects of the people's behaviour with mathematics. Questions around designing policies or interventions play a central role in economics. Thus they have transformed simultaneous equations into a powerful framework and associated set of methods for estimating causal effects. The is also a rich strand of work on using the assumptions that can be encoded in structural equation models, also known as functional causal models to discover the structure and direction of causal relationships - see for example \citet{mooij2016distinguishing}. Causal Bayesian networks \citep{Pearl2000} are a more recent development and arise from the addition of a fundamental assumption about the meaning of a link to Bayesian networks. They inherit and leverage the way Bayesian networks encode conditional independencies between variables to localise the impact of an intervention in a system in a way that allows formalisation of the conditions under which causal effects can be inferred from observational data.

However, the literature on causal inference techniques remains split between the different frameworks. Much of the recent work on estimating causal effects within machine learning, as well as widely used methodologies such as propensity scoring, are described using the counterfactual framework. Methods developed within economics, in particular instrumental variable based approaches, or those requiring parametric or functional assumptions, are often based around structural equation models. This makes it worthwhile for researchers interested in causality to develop an understanding of all these viewpoints.

In the next sections, we describe causal Bayesian networks, counterfactuals and structural equation models: the problems they allow us to solve, the assumptions they rely on and how they differ. By describing all three frameworks, how they relate to one-another, and when they can be viewed as equivalent, we will make it easier for researchers familiar with one framework to understand the others and to transfer ideas and techniques between them. In order to demonstrate the notation and formalisms each framework provides, we will use them to describe the following simple examples.

\begin{example}
\label{exm:ranomized_experiment}
Suppose a pharmaceutical company wants to assess the effectiveness of a new drug on recovery from a given illness. This is typically tested by taking a large group of representative patients and randomly assigning half of them to a treatment group (who receive the drug) and the other half to a control group (who receive a placebo). The goal is to determine the clinical impacts of the drug by comparing the differences between the outcomes for the two groups (in this case, simplified to only two outcomes - recovery or non-recovery). We will use the variable $X$ (1 = drug, 0 = placebo) to represent the treatment each person receives and $Y$ (1 = recover, 0 = not recover) to describe the outcome.
\end{example}

\begin{example}
\label{exm:adjusting}
Suppose we want to estimate the impact on high school graduation rates of compulsory preschool for all four year olds. We have a large cross-sectional data set on a group of twenty year olds that records if they attended preschool, if they graduated high school and their parents socio-economic status (SES). We will let $X\in \set{0,1}$ indicate if an individual attended preschool, $Y \in \set{0,1}$ indicate if they graduated high school and $Z \in \set{0,1}$ represent if they are from a low or high SES background respectively.\footnote{There has been substantial empirical work on the effectiveness of early childhood education including a landmark randomised trial, the Perry Preschool project, which ran from 1962-1967 \citep{weikart1970longitudinal}.}
\end{example}

\subsection{Causal Bayesian networks}

Causal Bayesian networks are an extension of Bayesian networks. A Bayesian network is a graphical way of representing how a distribution factorises. Any joint probability distribution can be factorised into a product of conditional probabilities. There are multiple valid factorisations, corresponding to permutations of variable ordering.

\eqn{
\label{eqn:cbn:joint_dist}
P(X_{1},X_{2},X_{3},...)=P(X_{1})P(X_{2}|X_{1})P(X_{3}|X_{1},X_{2})...
}

We can represent this graphically by drawing a network with a node for each variable and adding links from the variables on the right hand side to the variable on the left for each conditional probability distribution, see figure \ref{fig:bayesnet}. If the factorisation simplifies due to conditional independencies between variables, this is reflected by missing edges in the corresponding network. There are multiple valid Bayesian network representations for any probability distribution over more than one variable, see figure \ref{fig:bayesnet2} for an example.

\begin{figure}[ht]
\centering
\begin{tikzpicture}[->,>=stealth',shorten >=1pt,auto,node distance=1cm,
  thick,main node/.style={observed}]

\node[main node](1){$X_{1}$};
\node[main node, below left=of 1](2){$X_{2}$};
\node[main node, below right=of 1](3){$X_{3}$};

 \path[every node/.style={font=\sffamily\small}]
    (1) edge node {} (2)
    	edge node {} (3)
    (2) edge node {} (3);

\end{tikzpicture}
\caption{A general Bayesian network for the joint distribution over three variables. This network does not encode any conditional independencies between its variables and can thus represent any distribution over three variables.}
\label{fig:bayesnet}
\end{figure}
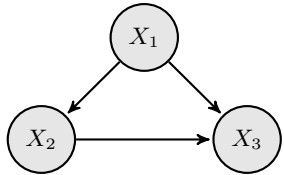

\begin{figure}
\centering
\begin{subfigure}[t]{0.15\textwidth}
\centering
\caption{}
\begin{tikzpicture}[->,>=stealth',shorten >=1pt,auto,node distance=1cm, thick,main node/.style={observed}]
\node[main node](1){$X_{1}$};
\node[main node, below=of 1](2){$X_{2}$};
\node[main node, below=of 2](3){$X_{3}$};
\path[every node/.style={font=\sffamily\small}]
    (1) edge (2)
    (2) edge (3);
\end{tikzpicture}
\end{subfigure}
\begin{subfigure}[t]{0.15\textwidth}
\centering
\caption{}
\begin{tikzpicture}[->,>=stealth',shorten >=1pt,auto,node distance=1cm, thick,main node/.style={observed}]
\node[main node](1){$X_{1}$};
\node[main node, below=of 1](2){$X_{2}$};
\node[main node, below=of 2](3){$X_{3}$};
\path[every node/.style={font=\sffamily\small}]
    (3) edge (2)
    (2) edge (1);
\end{tikzpicture}
\end{subfigure}
\begin{subfigure}[t]{0.3\textwidth}
\centering
\caption{}
\begin{tikzpicture}[->,>=stealth',shorten >=1pt,auto,node distance=1cm, thick,main node/.style={observed}]
\node[main node](1){$X_{2}$};
\node[main node, below left=of 1](2){$X_{1}$};
\node[main node, below right=of 1](3){$X_{3}$};
\path[every node/.style={font=\sffamily\small}]
    (1) edge (2) edge (3);
\end{tikzpicture}
\end{subfigure}
\begin{subfigure}[t]{0.3\textwidth}
\centering
\caption{}
\begin{tikzpicture}[->,>=stealth',shorten >=1pt,auto,node distance=1cm, thick,main node/.style={observed}]
\node[main node](1){$X_{2}$};
\node[main node, above left=of 1](2){$X_{1}$};
\node[main node, above right=of 1](3){$X_{3}$};
\path[every node/.style={font=\sffamily\small}]
    (2) edge (1)
    (3) edge (1)
    (2) edge (3);
\end{tikzpicture}
\end{subfigure}
\caption{Some valid Bayesian networks for a distribution $P$ over $(X_1,X_2,X_3)$ in which $X_3$ is conditionally independent of $X_1$ given $X_2$, denoted $X_3 \ci X_1 | X_2$. Graphs (a), (b) and (c) are all a \emph{perfect map} for $P$ as the graphical structure implies exactly the same set of independencies exhibited by the distribution. Graph (d), like figure \ref{fig:bayesnet} does not imply any conditional independencies, and is thus a valid (but not very useful) Bayesian network representation for any distribution over three variables.}
\label{fig:bayesnet2}
\end{figure}
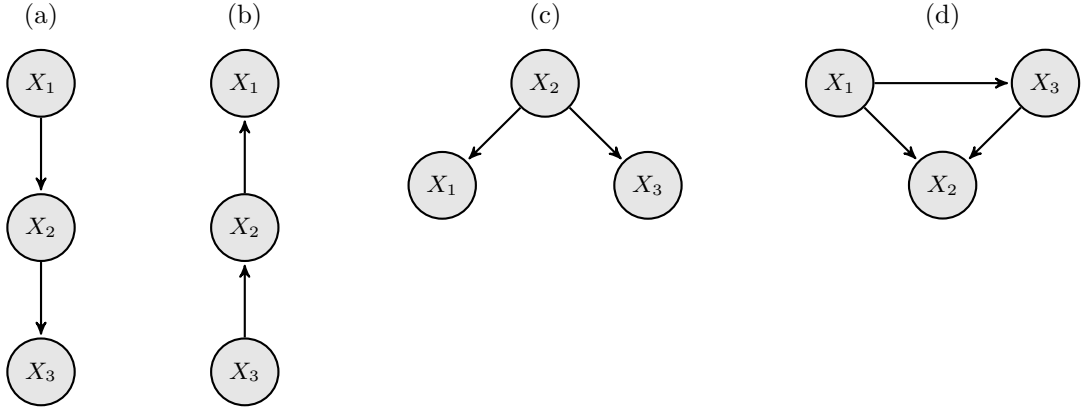

The statement that a given graph $G$ is a Bayesian network for a distribution $P$ tells us that the distribution can be factorised over the nodes and edges in the graph. There can be no missing edges in $G$ that do not correspond to conditional independencies in $P$, (the converse is not true: $G$ can have extra edges). If we let $\parents{X_{i}}$ represent the set of variables that are parents of the variable $X_{i}$ in $G$ then we can write the joint distribution as;

\eqn{
\P{X_{1},...,X_{N}} = \prod_{i = 1...N}\P{X_{i}|\parents{X_{i}}}
}

A causal Bayesian network is a Bayesian network in which a link $X_{i} \rightarrow X_{j}$, by definition, implies $X_{i}$ causes $X_{j}$. This means an intervention to change the value of $X_{i}$ can be expected to affect $X_{j}$, but interventions on $X_{j}$ will not affect $X_{i}$. We need some notation to describe interventions and represent distributions over variables in the network after an intervention. We use the do operator introduced by Pearl \citep{Pearl2000}.

\vspace{0.5cm}
\begin{definition}{The do-notation}
\begin{itemize}
\item $do(X=x)$ denotes an intervention that sets the random variable(s) $X$ to $x$.
\item $\P{Y|do(X)}$ is the distribution of $Y$ conditional on an \emph{intervention} that sets $X$. This notation is somewhat overloaded. It may be used to represent a probability distribution/mass function or a family of distribution functions depending on whether the variables are discrete or continuous and whether or not we are treating them as fixed. For example, it could represent
\begin{itemize}
\item the probability $\P{Y=1|do(X=x)}$ as a function of $x$,
\item the probability mass function for a discrete $Y$ : $\P{Y|do(X=x)}$,
\item the probability density function for a continuous  $Y$ : $f_Y(y|do(X=x))$,
\item a family of density/mass function for $Y$ parameterised by $x$.
\end{itemize}
Where the distinction is important and not clear from context we will use one of the more specific forms above.
\end{itemize}
\end{definition}

\vspace{0.5cm}
\begin{theorem}[Truncated product formula \citep{Pearl2000}]
\label{thm:truncated_prodcut}
If $G$ is a causal network for a distribution $P$ defined over variables $X_{1}...X_{N}$, then we can calculate the distribution after an intervention where we set $Z \subset X$ to $z$, denoted $do(Z=z)$ by dropping the terms for each of the variables in $Z$ from the factorisation given by the network. Let $\parents{X_i}$ denote the parents of the variable $X_i$ in $G$.

\begin{equation}
\label{eq:truncatedproduct}
\P{X_1...X_N|do(Z=z)} = \ind{Z = z}
  \prod_{X_i \notin Z}\P{X_{i}|\parents{X_i}}
\end{equation}
\end{theorem}

Theorem \ref{thm:truncated_prodcut} does not hold for standard Bayesian networks because there are multiple valid networks for the same distribution. The truncated product formula will give different results depending on the selected network. The result is possible with causal Bayesian networks because it follows directly from the assumption that the direction of the link indicates causality. In fact, from the interventionist viewpoint of causality, the truncated product formula defines what it means for a link to be causal.

\begin{figure}[ht]
\centering
\begin{tikzpicture}[->,shorten >=0pt,shorten <=0pt,node distance=2.5em,thick,node/.style={observedrect},lt/.style={latent}]
\node[node](2){$X \text{ (Treatment)}$};
\node[node, right=of 2](3){$Y \text{ (Outcome)}$};
\path[]
	(2) edge (3);
\end{tikzpicture}
\caption{Causal Bayesian network for example \ref{exm:ranomized_experiment}}
\label{fig:causal_network_example}
\end{figure}
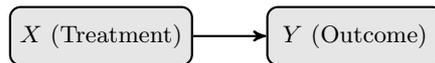

Returning to example \ref{exm:ranomized_experiment}, and phrasing our query in terms of interventions; what would the distribution of outcomes look like if everyone was treated $\P{Y|do(X=1)}$, relative to if no one was treated $\P{Y|do(X=0)}$? The treatment $X$ is a potential cause of $Y$, along with other unobserved variables, such as the age, gender and the disease subtype of the patient. Since $X$ is assigned via deliberate randomisation, it cannot be affected by any latent variables. The causal Bayesian network for this scenario is shown in figure \ref{fig:causal_network_example}. This network represents the (causal) factorisation  $\P{X,Y} = \P{X}\P{Y|X}$, so from equation (\ref{eq:truncatedproduct}), $\P{Y|do(X)} = \P{Y|X}$. In this example, the interventional distribution is equivalent to the observational one.

\begin{figure}
\center
\begin{tikzpicture}[->,>=stealth',shorten >=1pt,auto,node distance=1.2cm, thick,node/.style={observedrect},lt/.style={latent}]

\node[node](1){$Z \text{ (SES)}$};
\node[node, below left=of 1](2){$X \text{ (Pre-school)}$};
\node[node, below right=of 1](3){$Y \text{ (Graduated)}$};
\path[every node/.style={font=\sffamily\small}]
    (1) edge (2) edge (3)
    (2) edge (3);
\end{tikzpicture}
\caption{Causal Bayesian network for example \ref{exm:adjusting}}
\label{fig:causal_adjust}
\end{figure}
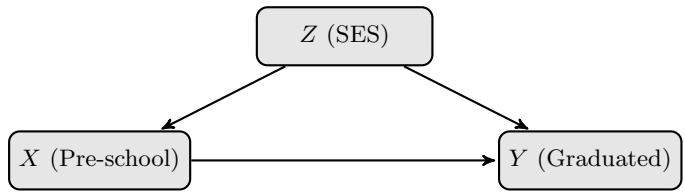

In example \ref{exm:adjusting} we are interested in $\P{Y|do(X=1)}$, the expected high-school graduation rate if we introduce universal preschool. We could compare it to outlawing preschool $\P{Y|do(X=0)}$ or the current status quo $\P{Y}$. It seems reasonable to assume that preschool attendance affects the likelihood of high school graduation \footnote{The effect does not have to be homogeneous, it may depend non-linearly on characteristics of the child, family and school.} and that parental socio-economic status would affect \emph{both} the likelihood of preschool attendance and high school graduation. If we assume that socio-economic status is the only such variable (nothing else affects both attendance \emph{and} graduation), we can represent this problem with the causal Bayesian network in figure \ref{fig:causal_adjust}. In this case, the interventional distribution is not equivalent to the observational one. If parents with high socio-economic status are more likely to send their children to preschool and these children are more likely to graduate high school regardless, comparing the graduation rates of those who attended preschool with those who did not will overstate the benefit of preschool. To obtain the interventional distribution we have to estimate the impact of preschool on high school graduation for each socio-economic level separately and then weight the results by the proportion of the population in that group,

\eqn{
\label{eqn:backdoor_example}
\P{Y|do(X=1)} = \sum_{z \in Z}\P{Y|X=1,Z}\P{Z}
}

We have seen from these two examples that the expression to estimate the causal effect of an intervention depends on the structure of the causal graph. There is a very powerful and general set of rules that specifies how we can transform observational distributions into interventional ones for a given graph structure. These rules are referred to as the Do-calculus \citep{Pearl2000}.

A causal Bayesian network represents much more information than a Bayesian network with identical structure. A causal network encodes all possible interventions that could be specified with the do-notation. For example, if the network in figure \ref{fig:causal_adjust} were an ordinary Bayesian network and all the variables were binary, the associated distribution could be described by seven parameters. The equivalent causal Bayesian network additionally represents the post-interventional distributions for six possible single variable interventions and twelve possible two variable interventions. Encoding all this information without the assumptions implicit in the causal Bayesian network would require an additional thirty parameters.\footnote{After each single variable intervention we have a distribution over two variables, which can be represented by three parameters. After each two variable intervention, we have a distribution over one variable which requires one parameter. This takes us to a total of $6\times3+12\times1 = 30$ additional parameters.}

Causal Bayesian networks are Bayesian networks, so results that apply to Bayesian networks carry directly across: the local Markov property states that variables are independent of their non-effects given their direct causes. The global Markov property and d-separation also hold in causal networks. D-separation, which characterises which conditional independencies must hold in any distribution that can be represented by a given Bayesian network $G$, is key to many important results and algorithms for causal inference.

\subsection{Limitations of causal Bayesian networks}
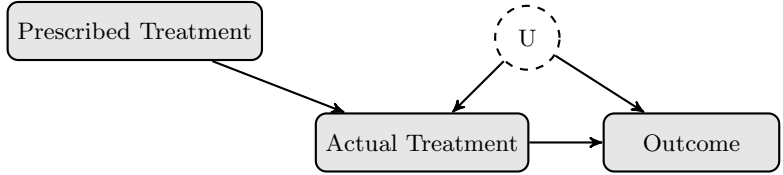
\begin{figure}[ht]
\centering
\begin{tikzpicture}[->,shorten >=0pt,shorten <=0pt,node distance=2.5em,thick,node/.style={observedrect},lt/.style={latent}]
\node[node](1){Prescribed Treatment};
\node[node, below right=of 1](2){Actual Treatment};
\node[node, right=of 2](3){Outcome};
\node[lt, above left=of 3](4){U};
\path[]
	(1) edge (2)
	(2) edge (3)
	(4) edge (3) edge (2);
\end{tikzpicture}
\caption{Randomised treatment with imperfect compliance}
\label{fig:randomized_imperfect_compliance}
\end{figure}

A number of criticisms have been levelled at this approach to modelling causality. One is that the definition of an intervention only in terms of setting the value of one or more variables is too precise and that any real world intervention will affect many variables in complex and non-deterministic ways \citep{rickles2009causality,cartwright2007hunting}. However, by augmenting the causal graph with additional variables that model how interventions may take effect, the deterministic do operator can model more complex interventions. For example, in the drug treatment case, we assumed that all subjects complied, taking the treatment or placebo as assigned by the experimenter. But, what if some people failed to take the prescribed treatment? We can model this within the framework of deterministic interventions by adding a node representing what they were prescribed (the intervention) which probabilistically influences the treatment they actually receive (figure \ref{fig:randomized_imperfect_compliance}). Note that the fact that we no longer directly assign the treatment opens the possibility that an unobserved latent variable could affect both the actual treatment taken and the outcome.

Another key issue with causal Bayesian networks is that they cannot handle cyclic dependencies between variables. Such feedback loops are common in real-life systems, for example the relationship between supply and demand in economics or predator and prey in ecology. We might regard the underlying causal mechanisms in these examples to be acyclic; the number of predators at one time influences the number of prey in the next period and so on. However, if our measurements of these variables must be aggregated over time periods that are longer than the scale at which these interactions occur, the result is a cyclical dependency. Even were we able to measure on shorter timescales, there might then not be sufficient data on each variable for inference. Such problems have mostly been studied within the dynamical systems literature, typically focusing on understanding the stationary or equilibrium state of the system and making very specific assumptions about functional form in order to make problems tractable. \citet{Poole2013} compare the equilibrium approach to reasoning about cyclic problems with structural equation models, which we discuss in section \ref{sec:SEM} and that can be seen as Bayesian causal networks with additional functional assumptions.

\subsection{Counterfactuals}
\label{sec:counterfactuals}

The Neyman-Rubin model \citep{Rubin1974,Rubin1978,Rosenbaum1983, Rubin2005,Rubin2008} defines causality in terms of potential outcomes, or counterfactuals. Counterfactuals are statements about imagined or alternate realities, are prevalent in everyday language and may play a role in the development of causal reasoning in humans \citep{Weisberg2013}. Causal effects are differences in counterfactual variables: what the difference is between what would have happened if we did one thing versus what would have happened if we did something else.

In example \ref{exm:ranomized_experiment}, the causal effect of the drug relative to placebo for person $i$ is the difference between what would have happened if they were given the drug, denoted $\cf{y_{i}}{1}$ versus what would have happened if they got the placebo, $\cf{y_{i}}{0}$. The fundamental problem of causal inference is that we can only observe one of these two outcomes, since a given person can only be treated or not treated. The problem can be resolved if, instead of people, there are units that can be assumed to be identical or that will revert exactly to their initial state some time after treatment. This type of assumption often holds to a good approximation in the natural sciences and explains why researchers in these fields are less concerned with causal theory.

Putting aside any estimates of individual causal effects, it is possible to learn something about the distributions under treatment or placebo. Let $\cf{Y}{1}$ be a random variable representing the potential outcome if treated. The distribution of $\cf{Y}{1}$ is the distribution of $Y$ if everyone was treated. Similarly $Y^{0}$ represents the potential outcome for the placebo. The difference between the probability of recovery, across the population, if everyone was treated and the probability of recovery if everyone received the placebo is $\P{\cf{Y}{1}}-\P{\cf{Y}{0}}$. We can estimate (from an experimental or observational study):
\begin{itemize}
\item $\P{Y=1|X=1}$, the probability that those who took the treatment will recover
\item $\P{Y=1|X=0}$, the probability that those who were \emph{not} treated will recover
\end{itemize}

Now, for those who took the treatment, the outcome \emph{had} they taken the treatment $\cf{Y}{1}$ is the same as the observed outcome. For those who did not take the treatment, the observed outcome is the same as the outcome \emph{had} they not taken the treatment. Equivalently stated:

\eq{
\P{Y^{0}|X=0}&= \P{Y|X=0}\\
\P{Y^{1}|X=1}&=\P{Y|X=1}
}

If we assume $X \ci Y^{0}$ and $X \ci Y^{1}$:

\eq{
\P{Y^{1}} &= \P{Y^{1}|X=1} = \P{Y|X=1} \\
\P{Y^{0}} &= \P{Y^{0}|X=0} = \P{Y|X=0}
}

This implies the counterfactual distributions are equivalent to the corresponding conditional distributions and, for a binary outcome $Y$, the causal effect is,

\eq{
\P{Y^{1}}-\P{Y^{0}} = \P{Y|X=1} - \P{Y|X=0}
}

The assumptions $X \ci Y^{1}$ and $X \ci Y^{0}$  are referred to as ignorability assumptions \citep{Rosenbaum1983}. They state that the treatment each person receives is independent of whether they would recover if treated and if they would recover if not treated. This is justified in example \ref{exm:ranomized_experiment} due to the randomisation of treatment assignment. In general the treatment assignment will not be independent of the potential outcomes. In example \ref{exm:adjusting}, the children from wealthy families could be more likely to attend preschool but also more likely to do better in school regardless, i.e
$X \nci \cf{Y}{0}$ and $X \nci \cf{Y}{1}$. A more general form of the ignorability assumption is to identify a set of variables $Z$ such that $X \ci Y^{1}|Z$ and $X \ci Y^{0}|Z$.

\vspace*{.3cm}
\begin{theorem}[Ignorability \citep{Rosenbaum1983, Pearl2000}] If $X \ci Y^{1}|Z$ and $X \ci Y^{0}|Z$,

\eqn{
\label{eqn:counterfactual1}
\P{\cf{Y}{1}} &= \sum_{z \in Z}\P{Y|X=1,Z}\P{Z}  \\
\label{eqn:counterfactual2}
\P{\cf{Y}{0}} &= \sum_{z \in Z}\P{Y|X=0,Z}\P{Z}
}
\end{theorem}

Assuming that within each socio-economic status level, attendance at preschool is independent of the likelihood of graduating high-school had a person attended, then the average rate of high-school graduation given a universal preschool program can be computed from equation \ref{eqn:counterfactual1}. Note, that this agrees with the weighted adjustment formula in equation \ref{eqn:backdoor_example}.

Another assumption introduced within the Neyman-Rubin causal framework is the Stable Unit Treatment Value Assumption (SUTVA) \citep{Rubin1978}. This is the assumption that the potential outcome for one individual (or unit) does not depend on the treatment assigned to another individual. As an example of a SUTVA violation, suppose disadvantaged four year olds were randomly assigned to attend preschool. The subsequent school results of children in the control group, who did not attend, could be boosted by the improved behaviour of those who did and who now share the classroom with them. SUTVA violations would manifest as a form of model misspecification in causal Bayesian networks.

There are objections to counterfactuals arising from the way they describe alternate universes that were never realised. In particular, statements involving joint distributions over counterfactual variables may not be able to be validated empirically \citet{Dawid2000}. One way of looking at counterfactuals is as a natural language short hand for describing highly specific interventions like those denoted by the do-notation. Rather than talking about the distribution of $Y$ given we intervene to set $X=x$ and hold everything else about the system constant we just say what would the distribution of $Y$ be had $X$ been $x$. This is certainly convenient, if rather imprecise. However, the ease with which we can make statements with counterfactuals that cannot be tested with empirical data warrants careful attention. It is important to be clear what assumptions are being made and whether or not they could be validated (at least in theory).

\subsection{Structural Equation models}
\label{sec:SEM}

Structural equation models (SEMs) describe a deterministic world, where some underlying mechanism or function determines the output of any process for a given input. The mechanism (but not the output) is assumed to be independent of what is fed into it. Uncertainties are not inherent but arise from unmeasured variables. Linear structural equation models have a long history for causal estimation \cite {Wright1921,Haavelmo1943}. More recently, they have been formalised, generalised to the non-linear setting and connected to developments in graphical models to provide a powerful causal framework \citep{Pearl2000}.

\begin{figure}[ht]
\centering
\begin{tikzpicture}[->,shorten >=0pt,shorten <=0pt,node distance=2.5em,thick,node/.style={observedrect},lt/.style={latent}]
\node[node](2){$X = f_x(\epsilon_x)$};
\node[node, right=of 2](3){$Y = f_y(X,\epsilon_y)$};
\node[lt, above=of 3](4){$\epsilon_y$};
\node[lt, above=of 2](5){$\epsilon_x$};
\path[]
	(2) edge (3)
	(4) edge (3)
	(5) edge (2);
\end{tikzpicture}
\caption{SEM for example \ref{exm:ranomized_experiment}}
\label{fig:sem_randomized_treatment}
\end{figure}
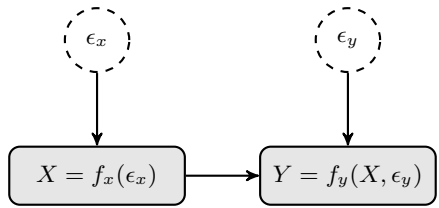

Mathematically, each variable is a deterministic function of its direct causes and a noise term that captures unmeasured variables. The noise terms are required to be mutually independent. If there is the possibility that an unmeasured variable influences more than one variable of interest in a study, it must be modelled explicitly as a latent variable. Structural equation models can be represented visually as a network. Each variable is a node and arrows are drawn from causes to their effects. Figure \ref{fig:sem_randomized_treatment} illustrates the SEM for example \ref{exm:ranomized_experiment}.

This model encodes the assumption that the outcome $y_{i}$ for an individual $i$ is caused solely by the treatment $x_{i}$ they receive and other factors $\epsilon_{y_{i}}$ that are independent of $X$. This is justifiable on the grounds that $X$ is random. The outcome of a coin flip for each patient should not be related to any of their characteristics (hidden or otherwise). Note that the causal graph in figure \ref{fig:sem_randomized_treatment} is identical to that of the Bayesian network for the same problem (figure \ref{fig:causal_network_example}). The latent variables $\epsilon_x$ and $\epsilon_y$ are not explicitly drawn in figure \ref{fig:causal_network_example} as they are captured by the probabilistic nature of the nodes in a Bayesian network.

Taking the \emph{action} $X=1$ corresponds to replacing the equation $X=f_x(\epsilon_x)$ with $X=1$. The function $f_y$ and distribution over $\epsilon_y$ does not change. This results in the interventional distribution, \footnote{We have assumed the variables are discrete only for notational convenience}

\eqn {
\P{Y=y|do(X=1)} = \sum_{\epsilon_y}\P{\epsilon_y}\ind{f_y(1,\epsilon_y)=y}
}

The observational distribution of $Y$ given $X$ is,

\eqn{
\P{Y=y|X=1} &= \sum_{\epsilon_x}\sum_{\epsilon_y}\P{\epsilon_x|X=1}\P{\epsilon_y|\epsilon_x}\ind{f_y(1,\epsilon_y)=y} \\
& = \sum_{\epsilon_y}\P{\epsilon_y}\ind{f_y(1,\epsilon_y)=y} \text{, as } \epsilon_x \ci \epsilon_y
}

The interventional distribution is the same as the observational one. The same argument applies to the intervention $do(X=0)$, and so the causal effect is simply the difference in observed outcomes as found via the causal Bayesian network and counterfactual approaches.

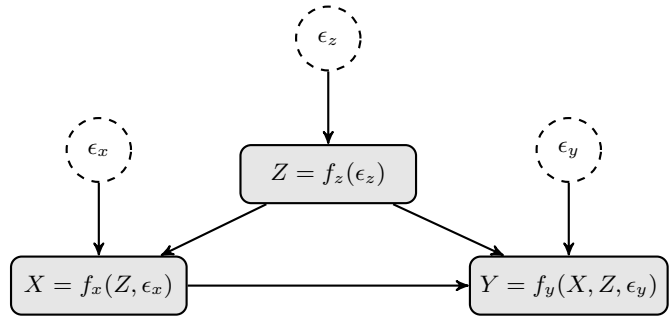
\begin{figure}[ht]
\centering
\begin{tikzpicture}[->,shorten >=0pt,shorten <=0pt,node distance=2.5em,thick,node/.style={observedrect},lt/.style={latent}]
\node[node](2){$X = f_x(Z,\epsilon_x)$};
\node[node,above right=of 2](6){$Z = f_z(\epsilon_z)$};
\node[lt, above=of 6](7){$\epsilon_z$};
\node[node, below right=of 6](3){$Y = f_y(X,Z,\epsilon_y)$};
\node[lt, above=of 3](4){$\epsilon_y$};
\node[lt, above=of 2](5){$\epsilon_x$};
\path[]
	(2) edge (3)
	(4) edge (3)
	(5) edge (2)
	(7) edge (6)
	(6) edge (2) edge (3);
\end{tikzpicture}
\caption{SEM for example \ref{exm:adjusting}}
\label{fig:sem:preschool}
\end{figure}

The SEM for example \ref{exm:adjusting} is shown in figure \ref{fig:sem:preschool}. Intervening to send all children to preschool replaces the equation $X = f_x(Z,\epsilon_x)$ with $X=1$, leaving all the other functions and distributions in the model unchanged.

\eqn{
\P{Y=y|do(X=1)} &= \sum_{z}\sum_{\epsilon_y}\P{z}\P{\epsilon_y}\ind{f_y(1,z,\epsilon_y)=y} \\
\label{eqn:sem:adjusting}
&=\sum_{z}\P{z}\underbrace{\sum_{\epsilon_y}\P{\epsilon_y}\ind{f_y(1,z,\epsilon_y)=y}}_{\P{Y=y|X=1,Z=z}}
}
Equation \ref{eqn:sem:adjusting} corresponds to equations \ref{eqn:backdoor_example} and \ref{eqn:counterfactual1}. It is not equivalent to the observational distribution given by:

\eqn{
\P{Y=y|X=1} = \sum_{z}\sum_{\epsilon_y}\P{z|X=1}\P{\epsilon_y}\ind{f_y(1,z,\epsilon_y)=y}
}

Structural equation models are generally applied with strong constraints on the functional form of the relationship between the variables and noise, which is typically assumed to be additive, $X_i = f_i(\cdot)+\epsilon_i$. A structural equation model with $N$ variables resembles a set of $N$ simultaneous equations, with each variable playing the role of the dependent (left hand side) variable in one equation. However a SEM is, by definition, more than a set of simultaneous equations. By declaring it to be structural, we are saying that it represents \emph{causal} assumptions about the relationships between variables. When visualised as a network, the absence of an arrow between two variables encodes the assumption that one does not cause the other. The similarity between the notation used to describe and analyse structural equation models and simultaneous equations, combined with a reluctance to make explicit statements about causality, has led to some confusion in the interpretation of SEMs \citep{heckman2015causal,Pearl2000}.

\subsection{Comparing and unifying the models}
\label{sec:unifying_causal_models}

Remarkably for models developed relatively independently in fields with very different approaches and problems, causal Bayesian networks, counterfactuals and structural equation models can be nicely unified for interventional queries (those that can be expressed with the do-notation) \citep{Pearl2000}. These queries, and the assumptions required to answer them, can be mapped between the frameworks in a straightforward way, allowing techniques developed within one framework to be immediately applied within another. If the network for a structural equation model is acyclic, that is if starting from any node and following edges in the direction of the arrows you cannot return to the starting point, then it implies a recursive factorisation of the joint distribution over its variables. In other words, the network is a causal Bayesian network. All of the results that apply to causal Bayesian networks also apply to acyclic structural equation models.  Taking an action that sets a variable to a specific value equates to replacing the equation for that variable with a constant. This corresponds to dropping a term in the factorisation and the truncated product formula (equation \ref{eq:truncatedproduct}). Thus, the interventional query $P(Y|do(X))$ is identical in these two frameworks. We can also connect this to counterfactuals via:

\begin{equation}
\begin{aligned}
&\P{Y^{0}} \equiv P(Y|do(X=0)) \\
&\P{Y^{1}} \equiv P(Y|do(X=1))
\end{aligned}
\end{equation}

The assumption $\epsilon_{X} \ci \epsilon_{Y}$, stated for our structural equation model, translates to $X \ci (Y^{0},Y^{1})$ in the language of counterfactuals. When discussing the counterfactual model, we made the slightly weaker assumption:

\begin{equation}
\label{eq:weakignore}
X \ci Y^{0} \text{ and } X \ci Y^{1}
\end{equation}

It is possible to relax the independence of errors assumption for SEMs to correspond exactly with the form of equation (\ref{eq:weakignore}) without losing any of the power provided by d-separation and graphical identification rules \citep{Richardson2013}. The correspondence between the models for interventional queries (those that can be phrased using the do-notation) makes it straightforward to combine key results and algorithms developed within any of these frameworks. For example, you can draw a causal graphical network to determine if a problem is identifiable and which variables should be adjusted for to obtain an unbiased causal estimate. Then use propensity scores \citep{Rosenbaum1983} to estimate the effect. If non-parametric assumptions are insufficient for identification or lead to overly large uncertainties, you can specify additional assumptions by phrasing your model in terms of structural equations. The frameworks do differ when it comes to causal queries that involve joint or nested counterfactuals and cannot be expressed with the do-notation. These types of queries arise in the study of mediation \citep{Pearl2014,Imai2010a,VanderWeele2011} and in legal decisions, particularly on issues such as discrimination \citep{Pearl2000}. The graphical approach to representing causal knowledge can be extended to cover these types of questions via Single World Intervention Graphs \citep{Richardson2013}, which explicitly represent counterfactual variables in the graph.

In practice, differences in focus and approach between the fields in which each model dominates eclipse the actual differences in the frameworks. The work on causal graphical models \citep{Pearl2000,Sprites2000} focuses on asymptotic, non-parametric estimation and rigorous theoretical foundations. The Neyman-Rubin framework builds on the understanding of randomised experiments and generalises to quasi-experimental and observational settings, with a particular focus on non-random assignment to treatment. Treatment variables are typically discrete (often binary). This research emphasises estimation of average causal effects and provides practical methods for estimation, in particular, propensity scores; a method to control for multiple variables in high dimensional settings with finite data \citep{Rosenbaum1983}. In economics, inferring causal effects from non-experimental data to support policy decisions is central to the field. Economists are often interested in more informative measures of the distribution of causal effects than the mean and make extensive use of structural equation models, generally with strong parametric assumptions \citep{Heckman2008}. The central approach to estimation is regression - which naturally handles continuous variables while discrete variables are typically encoded as indicator variables. In addition, the parametric structural equation models favoured in economics can be extended to analyse cyclic (otherwise referred to as non-recursive) models. However, these differences are not fundamental to the frameworks. Functional assumptions can be specified on the conditional distributions of (causal) Bayesian networks, counterfactuals can readily represent continuous treatments (eg $\cf{Y}{x}$), and structural equation models can represent complex non-linear relationships between both continuous and discrete variables.

\subsection{Granger causality}

A discussion of approaches to (observational) causal inference would not be complete without a mention of Granger causality, \citep{Granger1969}. The fundamental idea behind Granger causality is to leverage the assumption that the future does not cause the past to test the existence and direction of a causal link between two time series. The basic approach is to test, for a pair of time series variables $X$ and $Y$, if $Y_t \ci (X_{1},...,X_{t-1}) | (Y_{1},...,Y_{t-1})$ - that is if the history of $X$ helps to predict $Y$ given the history of $Y$. The original formulation considered only pairs of variables and linear causal relationships but recent work has generalised the key idea to multiple variables and non-linear relationships. Unlike the previous models we have discussed, Granger causality does not provide us with a means to specify our assumptions about the causal structure between variables. Rather it aims to infer the causal structure of a structural equation model from observational data - subject to some assumptions. In some sense, Granger causality is closer to standard prediction tasks in machine learning than causal analysis.

\vspace{2cm}
\bibliographystyle{plainnat}
{\small
\bibliography{references}
}

\end{document}